\documentclass[11pt,a4paper]{article}
\usepackage[hyperref]{acl2019}
\usepackage{times}
\usepackage{url}
\usepackage{latexsym}
\usepackage{amsmath,amsthm,amssymb}
\usepackage[para,online]{threeparttable}
\usepackage{graphicx}
\usepackage{booktabs}
\usepackage{algorithm}
\usepackage[noend]{algpseudocode}
\usepackage[title]{appendix}

\aclfinalcopy 
\def\aclpaperid{404} 

\usepackage{amsmath,amsfonts,bm}

\newcommand{\newterm}[1]{{\bf #1}}

\newcommand{\secref}[1]{\S\ref{#1}}

\def\eqref#1{equation~\ref{#1}}

\def\1{\bm{1}}

\def\vg{{\bm{g}}}
\def\vh{{\bm{h}}}

\def\vy{{\bm{y}}}

\DeclareMathAlphabet{\mathsfit}{\encodingdefault}{\sfdefault}{m}{sl}
\SetMathAlphabet{\mathsfit}{bold}{\encodingdefault}{\sfdefault}{bx}{n}

\def\gC{{\mathcal{C}}}
\def\gD{{\mathcal{D}}}

\def\gF{{\mathcal{F}}}
\def\gG{{\mathcal{G}}}

\def\gS{{\mathcal{S}}}
\def\gT{{\mathcal{T}}}

\def\sU{{\mathbb{U}}}

\def\sX{{\mathbb{X}}}

\newcommand{\manmoe}{\texttt{MAN-MoE}}
\newcommand{\man}{\texttt{MAN}}
\newcommand{\moe}{\texttt{MoE}}
\algnewcommand{\LeftComment}[1]{\State \(\triangleright\) #1}
\newcommand{\pluseq}{\mathrel{+}=}

\newcommand{\expe}{\mathop{{}\mathbb{E}}}

\title{Multi-Source Cross-Lingual Model Transfer: Learning What to Share}

\author{
Xilun Chen\textsuperscript{\textdagger}\thanks{ \ \ Most work was done while the first author was an intern at Microsoft Research.}
\And
Ahmed Hassan Awadallah\textsuperscript{\textdaggerdbl}
\And
Hany Hassan\textsuperscript{\textdaggerdbl}
\AND
Wei Wang\textsuperscript{\textdaggerdbl}
\And
Claire Cardie\textsuperscript{\textdagger}
\AND
\textnormal{\textsuperscript{\textdagger}Cornell University}\\
Ithaca, NY\\
\small\texttt{\{xlchen,cardie\}@cs.cornell.edu}\\
\And
\textnormal{\textsuperscript{\textdaggerdbl}Microsoft Research}\\
Redmond, WA\\
\small\texttt{\{hassanam,hanyh,Wei.Wang\}@microsoft.com}\\
}

\begin{document}

\maketitle

\begin{abstract}
Modern NLP applications have enjoyed a great boost utilizing neural networks models.
Such deep neural models, however, are not applicable to most human languages due to the lack of annotated training data for various NLP tasks.
Cross-lingual transfer learning (CLTL) is a viable method for building NLP models for a low-resource target language by leveraging labeled data from other (source) languages.
In this work, we focus on the multilingual transfer setting where training data in multiple source languages is leveraged to further boost target language performance.

Unlike most existing methods that rely only on language-invariant features for CLTL, our approach coherently utilizes both language-invariant and language-specific features \emph{at instance level}.
Our model leverages adversarial networks to learn language-invariant features, and mixture-of-experts models to dynamically exploit the similarity between the target language and each individual source language\footnote{The code is available at \url{https://github.com/microsoft/Multilingual-Model-Transfer}.}.
This enables our model to learn effectively what to share between various languages in the multilingual setup. Moreover, when coupled with unsupervised multilingual embeddings, our model can operate in a \newterm{zero-resource} setting where neither \emph{target language training data} nor \emph{cross-lingual resources} are available.
Our model achieves significant performance gains over prior art, as shown in an extensive set of experiments over multiple text classification and sequence tagging tasks including a large-scale industry dataset.

\end{abstract}

\section{Introduction}\label{sec:intro}
Recent advances in deep learning enabled a wide variety of NLP models to achieve impressive performance, thanks in part to the availability of large-scale annotated datasets.
However, such an advantage is not available to most of the world languages since many of them lack the the labeled data necessary for training deep neural nets for a variety of NLP tasks.
As it is prohibitive to obtain training data for all languages of interest, \emph{cross-lingual transfer learning} (CLTL) offers the possibility of learning models for a \emph{target language} using annotated data from other languages (\emph{source languages})~\citep{H01-1035}.
In this paper, we concentrate on the more challenging \emph{unsupervised} CLTL setting, where \emph{no} target language labeled data is used for training.\footnote{In contrast, supervised CLTL assumes the availability of annotations in the target language.}

Traditionally, most research on CLTL has been devoted to the standard \newterm{bilingual transfer} (BLTL) case where training data comes from a single source language.
In practice, however, it is often the case that we have labeled data in a few languages, and would like to be able to utilize all of the data when transferring to other languages.
Previous work~\cite{McDonald:2011:MTD:2145432.2145440} indeed showed that transferring from multiple source languages could result in significant performance improvement.
Therefore, in this work, we focus on the multi-source CLTL scenario, also known as \newterm{multilingual transfer learning} (MLTL), to further boost the target language performance.

One straightforward method employed in CLTL is weight sharing, namely directly applying the model trained on the source language to the target after mapping both languages to a common embedding space.
As shown in previous work~\citep{chen2016adan}, however, the distributions of the hidden feature vectors of samples from different languages extracted by the same neural net remain divergent, and hence weight sharing is not sufficient for learning a language-invariant feature space that generalizes well across languages.
As such, previous work has explored using \emph{language-adversarial training}~\citep{chen2016adan,D17-1302} to extract features that are invariant with respect to the shift in language, using only (non-parallel) unlabeled texts from each language. 

On the other hand, in the MLTL setting, where multiple source languages exist, language-adversarial training will only use, for model transfer, the features that are common among all source languages and the target, which may be too restrictive in many cases.
For example, when transferring from English, Spanish and \emph{Chinese} to German, language-adversarial training will retain only features that are invariant across all four languages, which can be too sparse to be informative.
Furthermore, the fact that German is more similar to English than to Chinese is neglected because the transferred model is unable to utilize features that are shared only between English and German.

To address these shortcomings, we propose a new MLTL model that not only exploits language-invariant features, but also allows the target language to dynamically and selectively leverage language-specific features through a probabilistic attention-style mixture of experts mechanism (see \secref{sec:model}). This allows our model to learn effectively what to share between various languages.
Another contribution of this paper is that, when combined with the recent unsupervised cross-lingual word embeddings~\cite{lample2018word,chen2018umwe}, our model is able to operate in a \newterm{zero-resource} setting where neither \emph{task-specific target language annotations} nor \emph{general-purpose cross-lingual resources} (e.g.~parallel corpora or machine translation (MT) systems) are available.
This is an advantage over many existing CLTL works, making our model more widely applicable to many lower-resource languages.

We evaluate our model on multiple MLTL tasks ranging from text classification to named entity recognition and semantic slot filling, including a real-world industry dataset. Our model beats all baseline models trained, like ours, without cross-lingual resources. More strikingly, in many cases, it can match or outperform state-of-the-art models that have access to strong cross-lingual supervision (e.g.~commercial MT systems).

\section{Related Work}\label{sec:relatedwork}

The diversity of human languages is a critical challenge for natural language processing.
In order to alleviate the need for obtaining annotated data for each task in each language, cross-lingual transfer learning (CLTL) has long been studied~\citep[][\textit{inter alia}]{H01-1035,10.1007/978-3-540-45175-4_13}.

For \emph{unsupervised} CLTL in particular, where no target language training data is available, most prior research investigates the \textbf{bilingual transfer} setting.
Traditionally, research focuses on \emph{resource-based} methods, where general-purpose cross-lingual resources such as MT systems or parallel corpora are utilized to replace task-specific annotated data~\cite{P09-1027,P10-1114}.
With the advent of deep learning, especially adversarial neural networks~\cite{NIPS2014_5423_gan,Ganin:2016:DTN:2946645.2946704}, progress has been made towards \emph{model-based} CLTL methods.
\citet{chen2016adan} propose language-adversarial training that does not directly depend on parallel corpora, but instead only requires a set of bilingual word embeddings (BWEs).

On the other hand,  the \textbf{multilingual transfer} setting, although less explored, has also been studied~\cite{McDonald:2011:MTD:2145432.2145440,naseem-etal-2012-selective,tackstrom-etal-2013-target,10.1007/978-3-319-05476-6_3,zhang-barzilay-2015-hierarchical,AAAI1612236}, showing improved performance compared to using labeled data from one source language as in bilingual transfer.

Another important direction for CLTL is to learn cross-lingual word representations~\cite{klementiev-titov-bhattarai:2012:PAPERS,zou-EtAl:2013:EMNLP,DBLP:journals/corr/MikolovLS13}.
Recently, there have been several notable work for learning fully unsupervised cross-lingual word embeddings, both for the bilingual~\citep{zhang-EtAl:2017:Long5,lample2018word,P18-1073} and multilingual case~\citep{chen2018umwe}.
These efforts pave the road for performing CLTL without cross-lingual resources.

Finally, a related field to MLTL is multi-source domain adaptation~\cite{NIPS2008_3550}, where most prior work relies on the learning of domain-invariant features~\cite{NIPS2018_8075,N18-1111}.
\newcite{ruder2019latent} propose a general framework for selective sharing between domains, but their method learns static weights at the \emph{task level}, while our model can dynamically select what to share at the instance level.
A very recent work~\cite{guo-shah-barzilay:2018:EMNLP} attempts to model the relation between the target domain and each source domain.
Our model combines the strengths of these methods and is able to simultaneously utilize both the domain-invariant and domain-specific features in a coherent way.
\section{Model}\label{sec:model}
One commonly adopted paradigm for neural cross-lingual transfer is the \emph{shared-private} model~\citep{NIPS2016_6254}, where the features are divided into two parts: \emph{shared} (language-invariant) features and \emph{private} (language-specific) features.
As mentioned before, the shared features are enforced to be language-invariant via language-adversarial training, by attempting to fool a language discriminator.
Furthermore, \citet{N18-1111} propose a generalized shared-private model for the multi-source setting, where a \emph{multinomial adversarial network} (\man{}) is adopted to extract common features shared by all source languages as well as the target.
\begin{figure}
    \centering
    \includegraphics[width=\linewidth]{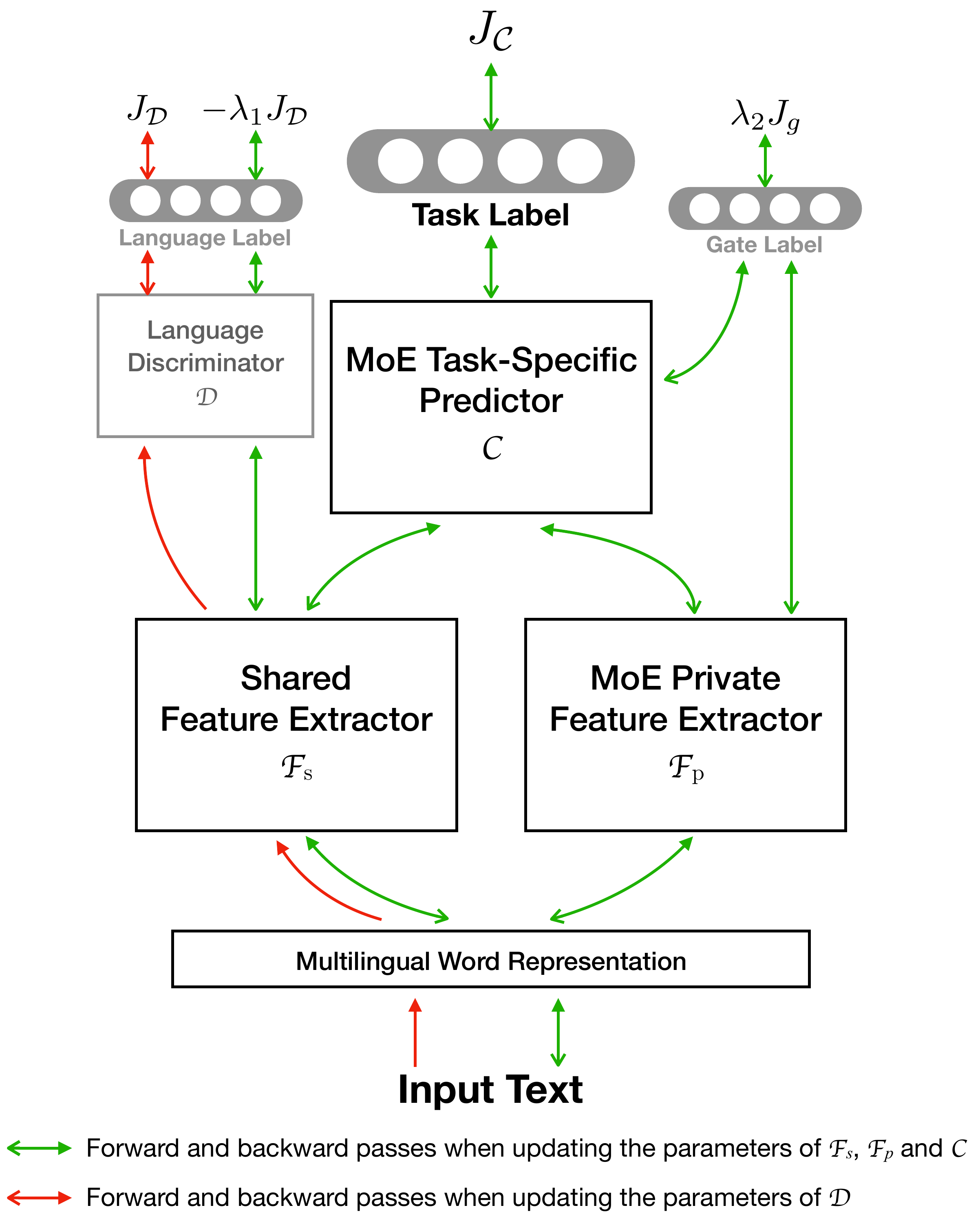}
    \caption{An overview of the \manmoe{} model.}
    \label{fig:manmoe}
\vspace{-3mm}
\end{figure}
On the other hand, the private features are learned by separate feature extractors, one for each source language, capturing the remaining features outside the shared ones.
During training, the labeled samples from a certain source language go through the corresponding private feature extractor for that particular language. At test time, there is no private feature extractor for the target language; only the shared features are used for cross-lingual transfer.

As mentioned in \secref{sec:intro}, using only the shared features for MLTL imposes an overly strong constraint and many useful features may be wiped out by adversarial training if they are shared only between the target language and a subset of source languages.
Therefore, we propose to use a mixture-of-experts (\moe{}) model~\citep{shazeer2017moe,N18-1032} to learn the private features.
The idea is to have a set of language expert networks, one per source language, each responsible for learning language-specific features for that source language during training.
However, instead of hard-switching between the experts, each sample uses a convex combination of all experts, dictated by an \emph{expert gate}.
Thus, at test time, the trained expert gate can decide the optimal expert weights for the unseen target language based on its similarity to the source languages.

Figure~\ref{fig:manmoe} shows an overview of our \manmoe{} model for multilingual model transfer.
The boxes illustrate various components of the \manmoe{} model~(\secref{sec:model_arch}), while the arrows depict the training flow (\secref{sec:training}).

\subsection{Model Architecture}\label{sec:model_arch}

Figure~\ref{fig:manmoe} portrays an abstract view of the \manmoe{} model with four components: the Multilingual Word Representation, the \man{} Shared Feature Extractor $\gF_s$ (together with the Language Discriminator $\gD$), the \moe{} Private Feature Extractor $\gF_p$, and finally the \moe{} Predictor $\gC$.
Based on the actual task (e.g.~sequence tagging, text classification, sequence to sequence, etc.), different architectures may be adopted, as explained below.

\noindent\textbf{Multilingual Word Representation}
embeds words from all languages into a single semantic space so that words with similar meanings are close to each other regardless of language.
In this work, we mainly rely on the MUSE embeddings~\citep{lample2018word}, which are trained in a fully unsupervised manner.
We map all other languages into English to obtain a multilingual embedding space.
However, in certain experiments, MUSE yields 0 accuracy on one or more language pairs~\citep{P18-1072}, in which case the VecMap embeddings~\citep{P17-1042} are used.
It uses \emph{identical strings} as supervision, which does not require parallel corpus or human annotations.
We further experiment with the recent unsupervised multilingual word embeddings~\cite{chen2018umwe}, which gives improved performance~(\secref{sec:exp:ner}).

In addition, for tasks where morphological features are important, one can add character-level word embeddings~\citep{DosSantos:2014:LCR:3044805.3045095} that captures sub-word information.
When character embeddings are used, we add a single CharCNN that is shared across all languages, and the final word representation is the concatenation of the word embedding and the char-level embedding. The CharCNN can then be trained end to end with the rest of the model.

\noindent\textbf{\man{} Shared Feature Extractor}
$\gF_s$ is a multinomial adversarial network~\citep{N18-1111}, which is an adversarial pair of a feature extractor (e.g.~LSTM or CNN) and a \emph{language discriminator} $\gD$.
$\gD$ is a text classifier~\citep{D14-1181} that takes the shared features (extracted by $\gF_s$) of an input sequence and predicts which language it comes from.
On the other hand, $\gF_s$ strives to fool $\gD$ so that it cannot identify the language of a sample.
The hypothesis is that if $\gD$ cannot recognize the language of the input, the shared features then do not contain language information and are hence language-invariant.
Note that $\gD$ is trained only using unlabeled texts, and can therefore be trained on all languages including the target language.

\begin{figure}
    \centering
    \hspace{-3mm}\includegraphics[width=\linewidth]{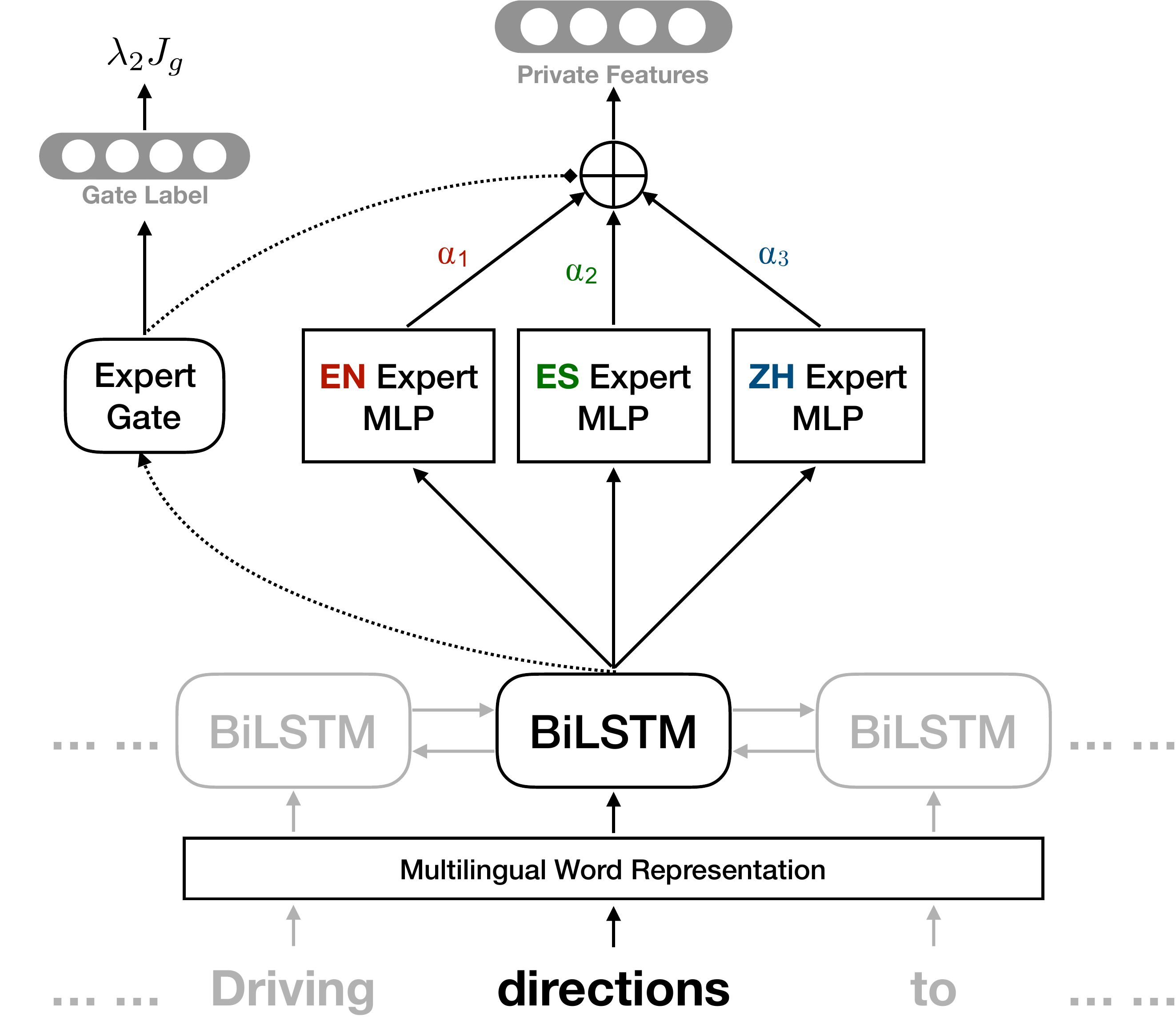}
    \caption{The \moe{} Private Feature Extractor $\gF_p$ with three source languages: English (EN), Spanish (ES), and Chinese (ZH).}
    \label{fig:fp}
\vspace{-3mm}
\end{figure}

\noindent\textbf{\moe{} Private Feature Extractor} $\gF_p$ is a key difference from previous work, shown in Figure~\ref{fig:fp}.
The figure shows the Mixture-of-Experts~\citep{shazeer2017moe} model with three source languages, English, Spanish, and Chinese.
$\gF_p$ has a shared BiLSTM at the bottom that extracts contextualized word representations for each token $w$ in the input sentence.
The LSTM hidden representation $\vh_w$ is then fed into the \moe{} module, where each source language has a separate expert network (a MLP).
In addition, the \emph{expert gate} $\gG$ is a linear transformation that takes $\vh_w$ as input and outputs a softmax score $\alpha_i$ for each expert.
The final private feature vector is a mixture of all expert outputs, dictated by the expert gate weights $\alpha$.

During training, the expert gate is trained to predict the language of a sample using the gate loss $J_g$, where the expert gate output $\alpha$ is treated as the softmax probability of the predicted languages.
In other words, the more accurate the language prediction is, the more the correct expert gets used.
Therefore, $J_g$ is used to encourage samples from a certain source language to use the correct expert, and each expert is hence learning language-specific features for that language.
As the BiLSTM is exposed to all source languages during training, the trained expert gate will be able to examine the hidden representation of a token to predict the optimal expert weights $\alpha$, even for \textbf{unseen} target languages at test time.
For instance, if a German test sample is similar to the English training samples, the trained expert gate will predict a higher $\alpha$ for the English expert, resulting in a heavier use of it in the final feature vector.
Therefore, even for the unforeseen target language (e.g.\ German), $\gF_p$ is able to dynamically determine what knowledge to use from each individual source language at a token level.

\noindent\textbf{\moe{} Task-Specific Predictor}
$\gC$ is the final module that make predictions for the end task, and may take different forms depending on the task.
For instance, for sequence tagging tasks, the shared and private features are first concatenated for each token, and then past through a \moe{} module similar to $\gF_p$ (as shown in Figure~\ref{fig:c} in the Appendix).
It is straightforward to adapt $\gC$ to work for other tasks.
For example, for text classification, a pooling layer such as dot-product attention~\citep{D15-1166} is added at the bottom to fuse token-level features into a single sentence feature vector.

$\gC$ first concatenates the shared and private features to form a single feature vector for each token.
It then has another \moe{} module that outputs a softmax probability over all labels for each token.
The idea is that it may be favorable to put different weights between the language-invariant and language-specific features for different target languages.
Again consider the example of English, German, Spanish and Chinese.
When transferring to Chinese from the other three, the source languages are similar to each other while all being rather distant from Chinese.
Therefore, the adversarially learned shared features might be more important in this case.
On the other hand, when transferring to German, which is much more similar to English than to Chinese, we might want to pay more attention to the \moe{} private features.
Therefore, we adopt a \moe{} module in $\gC$, which provides more flexibility than using a single MLP\footnote{We also experimented with an attention mechanism between the shared and private features, or a gating mechanism to modulate each feature channel, but got sub-optimal results.}.

\subsection{Model Training}\label{sec:training}
Denote the set of all $N$ source languages as $\gS$, where $|\gS|=N$.
Denote the target language as $\gT$, and let $\Delta=\gS \cup \gT$ be the set of all languages.
Denote the annotated corpus for a source language $l\in\gS$ as $\sX_l$, where $(x, y)\sim\sX_l$ is a sample drawn from $\sX_l$.
In addition, unlabeled data is required for all languages to facilitate the \man{} training.
We hence denote as $\sU_{l'}$ the unlabeled texts from a language $l'\in\Delta$.

\begin{algorithm}[t]
\small
\begin{algorithmic}[1]
\Require
labeled corpus $\mathbb{X}$; unlabeled corpus $\mathbb{U}$; Hyperpamameter $\lambda_1, \lambda_2 > 0$, $k \in \mathbb{N}$
\Repeat
\LeftComment{$\gD$ iterations}
\For{$diter = 1$ to $k$}
\State $l_\gD = 0$
\ForAll{$l \in \Delta$}\Comment{For all languages}
\State Sample a mini-batch $\bm{x} \sim \mathbb{U}_{l}$
\State $\bm{f}_s = \gF_s(\bm{x})$ \Comment{Shared features}
\State $l_\gD \pluseq L_\gD(\gD(\bm{f}_s); l)$ \Comment{$\gD$ loss}
\EndFor
\State Update $\gD$ parameters using $\nabla l_\gD$
\EndFor

\LeftComment{Main iteration}
\State $loss = 0$
\ForAll{$l \in \gS$}\Comment{For all source languages}
\State Sample a mini-batch $(\bm{x},\bm{y})  \sim \mathbb{X}_{l}$
\State $\bm{f}_s = \gF_s(\bm{x})$ \Comment{Shared features}
\State $\bm{f}_p, \vg_1 = \gF_p(\bm{x})$ \Comment{Private feat.\ \& gate outputs}
\State $\hat{\vy}, \vg_2 = \gC(\bm{f}_s, \bm{f}_p)$
\State $loss \pluseq L_\gC(\hat{\vy}; \bm{y}) + \lambda_2 (L_g(\vg_1; l) + L_g(\vg_2; l))$ 
\EndFor

\ForAll{$l \in \Delta$}\Comment{For all languages}
\State Sample a mini-batch $\bm{x} \sim \mathbb{U}_{l}$
\State $\bm{f}_s = \gF_s(\bm{x})$ \Comment{Shared features}
\State $loss \pluseq -\lambda_1\cdot L_{\gD}(\gD(\bm{f}_s); l)$ \Comment{Confuse $\gD$}
\EndFor
\State Update $\gF_s$, $\gF_p$, $\gC$ parameters using $\nabla loss$
\Until{convergence}

\end{algorithmic}
\caption{\manmoe{} Training}
\label{alg:training}
\end{algorithm}

The overall training flow of variant components is illustrated in Figure~\ref{fig:manmoe}, while the training algorithm is depicted in Algorithm~\ref{alg:training}.
Similar to \man{}, there are two separate optimizers to train \manmoe{}, one updating the parameters of $\gD$ (red arrows), while the other updating the parameters of all other modules (green arrows).
In Algorithm~\ref{alg:training}, $L_\gC$, $L_\gD$ and $L_g$ are the loss functions for the predictor $\gC$, the language discriminator $\gD$, and the expert gates in $\gF_p$ and $\gC$, respectively.

In practice, we adopt the NLL loss for $L_\gC$ for text classification, and token-level NLL loss for sequence tagging:
\begin{align}
L^{NLL}(\hat{y}; y) &= -\log P(\hat{y}=y) \label{eqn:nll_loss}\\
L^{T\text{-}NLL}(\hat{\vy}; \vy) &= -\log P(\hat{\vy} = \vy) \nonumber\\
&= -\sum_i \log P(\hat{y_i}=y_i) \label{eqn:token_nll_loss}
\end{align}
where $y$ is a scalar class label, and $\vy$ is a vector of token labels.
$L_\gC$ is hence interpreted as the \emph{negative log-likelihood} of predicting the correct task label.
Similarly, $\gD$ adopts the NLL loss in (\ref{eqn:nll_loss}) for predicting the correct language of a sample.
Finally, the expert gates $\gG$ use token-level NLL loss in (\ref{eqn:token_nll_loss}), which translates to the negative log-likelihood of using the correct language expert for each token in a sample.

Therefore, the objectives that $\gC$, $\gD$ and $\gG$ minimize are, respectively:
\begin{align}
    J_\gC &= \sum_{l\in\gS} \expe_{(x,y)\in\sX_l} \left[ L_\gC\left(\gC(\gF_s(x), \gF_p(x)); y\right)\right] 
    \label{eqn:j_c}\\
    J_\gD &= \sum_{l\in\Delta} \expe_{x\in\sU_l} \left[ L_\gD(\gD(\gF_s(x)); l)\right]
    \label{eqn:j_d}\\
    J_\gG &= \sum_{l\in\gS} \expe_{x\in\sX_l} \left[\sum_{w\in x} L_\gG(\gG(h_w); l)\right]
    \label{eqn:j_g}
\end{align}
where $\vh_w$ in (\ref{eqn:j_g}) is the BiLSTM hidden representation in $\gF_p$ as shown in Figure~\ref{fig:fp}.
In addition, note that $\gD$ is trained using unlabeled corpora over all languages ($\Delta$), while the training of $\gF_p$ and $\gC$ (and hence $\gG$) only take place on source languages ($\gS$).
Finally, the overall objective function is:
\begin{equation}
    J = J_\gC - \lambda_1 J_\gD + \lambda_2 \left(J_\gG^{(1)} + J_\gG^{(2)}\right)
    \label{eqn:j}
\end{equation}
where $J_\gG^{(1)}$ and $J_\gG^{(2)}$ are the two expert gates in $\gF_p$ and $\gC$, respectively.
More implementation details can be found in Appendix~\ref{sec:implementation}.
\section{Experiments}\label{sec:exp}

\begin{table*}
\centering
\small
\begin{tabular}{@{\hspace{0.43em}}l@{\hspace{0.43em}}@{\hspace{0.43em}}r@{\hspace{0.43em}}@{\hspace{0.43em}}r@{\hspace{0.43em}}@{\hspace{0.43em}}r@{\hspace{0.43em}}@{\hspace{0.43em}}r@{\hspace{0.43em}}@{\hspace{0.43em}}r@{\hspace{0.43em}}@{\hspace{0.43em}}r@{\hspace{0.43em}}@{\hspace{0.43em}}r@{\hspace{0.43em}}@{\hspace{0.43em}}r@{\hspace{0.43em}}@{\hspace{0.43em}}r@{\hspace{0.43em}}@{\hspace{0.43em}}r@{\hspace{0.43em}}@{\hspace{0.43em}}r@{\hspace{0.43em}}@{\hspace{0.43em}}r@{\hspace{0.43em}}@{\hspace{0.43em}}r@{\hspace{0.43em}}@{\hspace{0.43em}}r@{\hspace{0.43em}}@{\hspace{0.43em}}r@{\hspace{0.43em}}@{\hspace{0.43em}}r@{\hspace{0.43em}}}
\toprule
&   \multicolumn{3}{c}{English} &&   \multicolumn{3}{c}{German}   &&   \multicolumn{3}{c}{Spanish}  &&  \multicolumn{3}{c}{Chinese} &  \\
\cmidrule{2-4}\cmidrule{6-8}\cmidrule{10-12}\cmidrule{14-16}
Domain & \#Train & \#Dev & \#Test && \#Train & \#Dev & \#Test && \#Train & \#Dev & \#Test && \#Train & \#Dev & \#Test & \#Slot \\
\midrule
Navigation & 311045 & 23480 & 36625 && 13356 & 1599 & 2014 && 13862 & 1497 & 1986 && 7472 & 1114 & 1173 & 8 \\
Calendar & 64010 & 5946 & 8260 && 8261 & 1084 & 1366 && 6706 & 926 & 1081 && 2056 & 309 & 390 & 4 \\
Files & 30339 & 2058 & 5355 && 3005 & 451 & 480 && 6082 & 843 & 970 && 1289 & 256 & 215 & 5 \\
\specialrule{\heavyrulewidth}{\aboverulesep}{\belowrulesep}
Domain & \multicolumn{16}{c}{Examples} \\
\midrule
Navigation & \multicolumn{16}{l}{$[\it Driving]_{transportation\_type}$ directions to $[\it Walmart]_{place\_name}$ in $[\it New\ York]_{location}$.} \\
Calendar & \multicolumn{16}{l}{Add $[\it school\ meeting]_{title}$ to my calendar on $[\it Monday]_{start\_date}$ at $[\it noon]_{start\_time}$.} \\
Files & \multicolumn{16}{l}{Search for $[\it notes]_{data\_type}$ with $[\it grocery\ list]_{keyword}$.}\\
\bottomrule
\end{tabular}
\caption{Statistics for the Multilingual Semantic Slot Filling dataset with examples from each domain.}
\label{tab:slotfill_stats}
\vspace{-3mm}
\end{table*}

In this section, we present an extensive set of experiments across three datasets.
The first experiment is on a real-world multilingual slot filling (sequence tagging) dataset, where the data is used in a commercial personal virtual assistant.
In addition, we conduct experiments on two public academic datasets, namely the CoNLL multilingual named entity recognition (sequence tagging) dataset~\citep{W02-2024,W03-0419}, and the multilingual Amazon reviews (text classification) dataset~\citep{P10-1114}.

\subsection{Cross-Lingual Semantic Slot Filling}\label{sec:exp:slotfill}

As shown in Table~\ref{tab:slotfill_stats}, we collect data for four languages: English, German, Spanish, and Chinese, over three domains: Navigation, Calendar, and Files.
Each domain has a set of pre-determined slots (the slots are the same across languages), and the user utterances in each language and domain are annotated by crowd workers with the correct slots (see the examples in Table~\ref{tab:slotfill_stats}).
We employ the standard BIO tagging scheme to formulate the slot filling problem as a sequence tagging task.

For each domain and language, the data is divided into a training, a validation, and a test set, with the number of samples in each split shown in Table~\ref{tab:slotfill_stats}.
In our experiments, we treat each domain as a separate experiment, and consider each of German, Spanish and Chinese as the target language while the remaining three being source languages, which results in a total of 9 experiments.

\begin{table*}
\centering
\small
\begin{tabular}{@{\hspace{0.5em}}l@{\hspace{0.5em}}@{\hspace{0.5em}}c@{\hspace{0.5em}}@{\hspace{0.5em}}c@{\hspace{0.5em}}@{\hspace{0.5em}}c@{\hspace{0.5em}}@{\hspace{0.5em}}c@{\hspace{0.5em}}@{\hspace{0.5em}}c@{\hspace{0.5em}}@{\hspace{0.5em}}c@{\hspace{0.5em}}@{\hspace{0.5em}}c@{\hspace{0.5em}}@{\hspace{0.5em}}c@{\hspace{0.5em}}@{\hspace{0.5em}}c@{\hspace{0.5em}}@{\hspace{0.5em}}c@{\hspace{0.5em}}@{\hspace{0.5em}}c@{\hspace{0.5em}}@{\hspace{0.5em}}c@{\hspace{0.5em}}@{\hspace{0.5em}}c@{\hspace{0.5em}}@{\hspace{0.5em}}c@{\hspace{0.5em}}}
    \toprule
    &   \multicolumn{4}{c}{German}   &&   \multicolumn{4}{c}{Spanish}  &&  \multicolumn{4}{c}{Chinese} \\
    \cmidrule{2-5}\cmidrule{7-10}\cmidrule{12-15}
    Domain & Navi. & Cal. & Files & \emph{avg.} && Navi. & Cal. & Files & \emph{avg.} && Navi. & Cal. & Files & \emph{avg.} \\
    \midrule
    \multicolumn{15}{l}{\emph{Methods with cross-lingual resources}}\\
    MT (train-on-trans.)  & \underline{59.95} & \underline{63.53} & 38.68 & \underline{54.05} &  & \underline{64.37} & \bf \underline{59.93} & \underline{67.55} & \bf \underline{63.95} &  & \bf \underline{60.56} & \bf \underline{66.49} & 61.01 & \bf \underline{62.69} \\
    MT (test-on-trans.)  & 54.49 & 51.74 & \underline{55.87} & 54.03 &  & 52.13 & 58.10 & 55.00 & 55.08 &  & 54.23 & 22.71 & \bf \underline{64.01} & 46.98 \\ 
    \midrule
    \multicolumn{15}{l}{\emph{Methods without cross-lingual resources}}\\
    BWE (1-to-1)  & 57.53 & 58.28 & 35.73 & 50.51 &  & 62.54 & 44.44 & 57.56 & 54.85 &  & 17.62 & 22.48 & 21.32 & 20.47 \\ 
    BWE (3-to-1)  & 61.03 & 67.66 & 51.30 & 60.00 &  & 63.74 & 45.10 & 64.47 & 57.77 &  & 20.91 & 13.70 & 28.47 & 21.03 \\
    \man{}  & 59.07 & 60.24 & 39.35 & 52.89 &  & 58.86 & 37.90 & 46.75 & 47.84 &  & \underline{34.45} & 13.53 & 40.63 & 29.54 \\ 
    \manmoe{}  & \bf \underline{62.73} & \bf \underline{75.13} & \bf \underline{59.19} & \bf \underline{65.68} &  & \bf \underline{66.57} & \underline{50.21} & \bf \underline{70.91} & \underline{62.56} &  & 34.18 & \underline{29.36} & \underline{41.70} & \underline{35.08} \\ 
    \bottomrule
\end{tabular}
\caption{F1 scores on the Multilingual Semantic Slot Filling dataset. The highest performance is in bold; the highest performance within method group (with vs.\ without cross-lingual resources) is underlined (\emph{sic passim}).}
\label{tab:slotfill_results}
\end{table*}

\subsubsection{Results}

In Table~\ref{tab:slotfill_results}, we report the performance of \manmoe{} compared to a number of baseline systems.
All systems adopt the same base architecture, which is a multi-layer BiLSTM sequence tagger~\citep{irsoy-drnt} with a token-level MLP on top (no CRFs were used).

\noindent\newterm{MT baselines} employ machine translation (MT) for cross-lingual transfer.
In particular, the \emph{train-on-trans(lation)} method translates the entire English training set into each target language which are in turn used to train a supervised system on the target language.
On the other hand, the \emph{test-on-trans(lation)} method trains an English sequence tagger, and utilizes MT to translate the test set of each target language into English in order to make predictions.
In this work, we adopt the Microsoft Translator\footnote{\urlstyle{rm}\url{https://azure.microsoft.com/en-us/services/cognitive-services/translator-text-api/}}, a strong commercial MT system.
Note that for a MT system to work for sequence tagging tasks, \emph{word alignment} information must be available, in order to project word-level annotations across languages.
This rules out many MT systems such as Google Translate since they do not provide word alignment information through their APIs.

\begin{table*}
\centering
\small
\begin{tabular}{@{\hspace{0.6em}}l@{\hspace{0.6em}}@{\hspace{0.6em}}c@{\hspace{0.6em}}@{\hspace{0.6em}}c@{\hspace{0.6em}}@{\hspace{0.6em}}c@{\hspace{0.6em}}@{\hspace{0.6em}}c@{\hspace{0.6em}}@{\hspace{0.6em}}c@{\hspace{0.6em}}@{\hspace{0.6em}}c@{\hspace{0.6em}}@{\hspace{0.6em}}c@{\hspace{0.6em}}@{\hspace{0.6em}}c@{\hspace{0.6em}}@{\hspace{0.6em}}c@{\hspace{0.6em}}@{\hspace{0.6em}}c@{\hspace{0.6em}}@{\hspace{0.6em}}c@{\hspace{0.6em}}@{\hspace{0.6em}}c@{\hspace{0.6em}}@{\hspace{0.6em}}c@{\hspace{0.6em}}@{\hspace{0.6em}}c@{\hspace{0.6em}}}
    \toprule
    &   \multicolumn{4}{c}{German}   &&   \multicolumn{4}{c}{Spanish}  &&  \multicolumn{4}{c}{Chinese} \\
    \cmidrule{2-5}\cmidrule{7-10}\cmidrule{12-15}
    Domain & Navi. & Cal. & Files & \emph{avg} && Navi. & Cal. & Files & \emph{avg} && Navi. & Cal. & Files & \emph{avg} \\
    \midrule
    \manmoe{}  & 62.73 & 75.13 & \bf 59.19 & \bf 65.68 &  & \bf 66.57 & \bf 50.21 & \bf 70.91 & \bf 62.56 &  & 34.18 & \bf 29.36 & 41.70 & \bf 35.08 \\
    - $\gC$ \moe{}  & \bf 63.42 & \bf 76.68 & 55.68 & 65.26 &  & 65.50 & 47.51 & 69.67 & 60.89 &  & 27.71 & 21.75 & \bf 41.77 & 30.41 \\
    - $\gF_p$ \moe{}  & 58.33 & 48.85 & 37.35 & 48.18 &  & 58.99 & 36.67 & 48.39 & 48.02 &  & \bf 39.61 & 14.64 & 38.08 & 30.78 \\
    - both \moe{}  & 59.07 & 60.24 & 39.35 & 52.89 &  & 58.86 & 37.90 & 46.75 & 47.84 &  & 34.45 & 13.53 & 40.63 & 29.54 \\
    - \man{}  & 60.64 & 67.69 & 55.10 & 61.14 &  & 65.38 & 46.71 & 68.25 & 60.11 &  & 18.43 & 10.82 & 28.90 & 19.38 \\
    \bottomrule
\end{tabular}
\caption{Ablation (w.r.t.\ \manmoe{}) results on the Multilingual Semantic Slot Filling dataset.}
\label{tab:ablation}
\vspace{-3mm}
\end{table*}

\noindent\newterm{BWE baselines} rely on Bilingual Word Embeddings (BWEs) and weight sharing for CLTL.
Namely, the sequence tagger trained on the source language(s) are directly applied to the target language, in hopes that the BWEs could bridge the language gap.
This simple method has been shown to yield strong results in recent work~\citep{46604}.
The MUSE~\citep{lample2018word} BWEs are used by all systems in this experiment.
\emph{1-to-1} indicates that we are only transferring from English, while \emph{3-to-1} means the training data from all other three languages are leveraged.\footnote{\man{} and \manmoe{} results are always 3-to-1.}

The final baseline is the \man{} model~\citep{N18-1111}, presented before our \manmoe{} approach.
As shown in Table~\ref{tab:slotfill_results}, \manmoe{} substantially outperforms all baseline systems that do not employ cross-lingual supervision on almost all domains and languages.
Another interesting observation is that \man{} performs strongly on Chinese while being much worse on German and Spanish compared to the BWE baseline.
This corroborates our hypothesis that \man{} only leverages features that are invariant across \emph{all} languages for CLTL, and it learns such features better than weight sharing.
Therefore, when transferring to German or Spanish, which is similar to a subset of source languages, the performance of \man{} degrades significantly. 
On the other hand, when Chinese serves as the target language, where all source languages are rather distant from it, \man{} has its merit in extracting language-invariant features that could generalize to Chinese.
With \manmoe{}, however, this trade-off between close and distant language pairs is well addressed by the combination of \man{} and \moe{}.
By utilizing both language-invariant and language-specific features for transfer, \manmoe{} outperforms all cross-lingually unsupervised baselines on all languages.

Furthermore, even when compared with the MT baseline, which has access to hundreds of millions of parallel sentences, \manmoe{} performs competitively on German and Spanish.
It even significantly beats both MT systems on German as MT sometimes fails to provide accurate word alignment for German.
On Chinese, where the unsupervised BWEs are much less accurate (BWE baselines only achieve 20\% F1), \manmoe{} is able to greatly improve over the BWE and \man{} baselines and shows promising results for zero-resource CLTL even between distant language pairs.

\subsubsection{Feature Ablation}\label{sec:exp:ablation}

In this section, we take a closer look at the various modules of \manmoe{} and their impacts on performance (Table~\ref{tab:ablation}).
When the \moe{} in $\gC$ is removed, moderate decrease is observed on all languages.
The performance degrades the most on Chinese, suggesting that using a single MLP in $\gC$ is not ideal when the target language is not similar to the sources.
When removing the private \moe{}, the \moe{} in $\gC$ no longer makes much sense as $\gC$ only has access to the shared features, and the performance is even slightly worse than removing both \moe{}s.
With both \moe{} modules removed, it reduces to the \man{} model, and we see a significant drop on German and Spanish.
Finally, when removing \man{} while keeping \moe{}, where the shared features are simply learned via weight-sharing, we see a slight drop on German and Spanish, but a rather great one on Chinese.
The ablation results support our hypotheses and validate the merit of \manmoe{}.

\subsection{Cross-Lingual Named Entity Recognition}\label{sec:exp:ner}

In this section, we present experiments on the CoNLL 2002 \& 2003 multilingual named entity recognition (NER) dataset~\cite{W02-2024,W03-0419}, with four languages: English, German, Spanish and Dutch.
The task is also formulated as a sequence tagging problem, with four types of tags: PER, LOC, ORG, and MISC.

The results are summarized in Table~\ref{tab:ner_results}.
We observe that using only word embeddings does not yield satisfactory results, since the out-of-vocabulary problem is rather severe, and morphological features such as capitalization is crucial for NER.
We hence add character-level word embeddings for this task (\secref{sec:model_arch}) to capture subword features and alleviate the OOV problem.
For German, however, all nouns are capitalized, and the capitalization features learned on the other three languages would lead to poor results.
Therefore, for German only, we lowercase all characters in systems that adopt CharCNN.

\begin{table}
\begin{threeparttable}
    \centering
    \small
    \begin{tabular}{@{\hspace{0.4em}}l@{\hspace{0.4em}}@{\hspace{0.4em}}c@{\hspace{0.4em}}@{\hspace{0.4em}}c@{\hspace{0.4em}}@{\hspace{0.4em}}c@{\hspace{0.4em}}@{\hspace{0.4em}}c@{\hspace{0.4em}}}
        \toprule
        Target Language &   de  &   es & nl & avg \\
        \midrule
        \multicolumn{5}{l}{\emph{Methods with cross-lingual resources}}\\
        \citet{N12-1052} & 40.4 & 59.3 & 58.4 & 52.7 \\
        \citet{Nothman:2013:LMN:2405838.2405915} & 55.8 & 61.0 & 64.0 & 60.3 \\
        \citet{K16-1022} & 48.1 & 60.6 & 61.6 & 56.8 \\
        \citet{P17-1135} & \bf \underline{58.5} & 65.1 & \underline{65.4} & \underline{63.0} \\
        \citet{D17-1269} & 57.5 & \underline{66.0} & 64.5 & 62.3 \\
        \midrule
        \multicolumn{5}{l}{\emph{Methods without cross-lingual resources}}\\
        \manmoe{} & 55.1 & 59.5 & 61.8 & 58.8 \\
        BWE+CharCNN (1-to-1) & 51.5 & 61.0 & 67.3 & 60.0 \\
        BWE+CharCNN (3-to-1) & 55.8 & 70.4 & 69.8 & 65.3 \\
        \citet{D18-1034}\tnote{*} & \underline{56.9} & 71.0 & 71.3 & 66.4 \\
        \manmoe{}+CharCNN & 56.7 & 71.0 & 70.9 & 66.2 \\
        \manmoe{}+CharCNN+UMWE & 56.0 & \bf \underline{73.5} & \bf \underline{72.4} & \bf \underline{67.3} \\
        \bottomrule
    \end{tabular}
    \begin{tablenotes}
    \noindent\item[*] Contemporaneous work
    \end{tablenotes}
    \caption{F1 scores for the CoNLL NER dataset on German (de), Spanish (es) and Dutch (nl).}
    \label{tab:ner_results}
    \vspace{-3mm}
\end{threeparttable}
\end{table}

Table~\ref{tab:ner_results} also shows the performance of several state-of-the-art models in the literature\footnote{We also experimented with the MT baselines, but it often failed to produce word alignment, resulting in many empty predictions. The MT baselines attain only a F1 score of ${\sim}30\%$, and were thus excluded for comparison.}.
Note that most of these systems are specifically designed for the NER task, and exploit many task-specific resources, such as multilingual gazetteers, or metadata in Freebase or Wikipedia (such as entity categories).
Among these, \citet{N12-1052} rely on parallel corpora to learn cross-lingual word clusters that serve as features.
\citet{Nothman:2013:LMN:2405838.2405915,K16-1022} both leverage information in external knowledge bases such as Wikipedia to learn useful features for cross-lingual NER.
\citet{P17-1135} employ noisy parallel corpora (aligned sentence pairs, but not always translations) and bilingual dictionaries (5k words for each language pair) for model transfer.
They further add external features such as entity types learned from Wikipedia for improved performance.
Finally, \citet{D17-1269} propose a multi-source framework that utilizes large cross-lingual lexica.
Despite using none of these resources, general or task-specific, \manmoe{} nonetheless outperforms all these methods.
The only exception is German, where task-specific resources remain helpful due to its unique capitalization rules and high OOV rate.

\begin{table*}
\centering
\begin{threeparttable}
\small
\begin{tabular}{@{\hspace{0.5em}}l@{\hspace{0.5em}}@{\hspace{0.5em}}c@{\hspace{0.5em}}@{\hspace{0.5em}}c@{\hspace{0.5em}}@{\hspace{0.5em}}c@{\hspace{0.5em}}@{\hspace{0.5em}}c@{\hspace{0.5em}}@{\hspace{0.5em}}c@{\hspace{0.5em}}@{\hspace{0.5em}}c@{\hspace{0.5em}}@{\hspace{0.5em}}c@{\hspace{0.5em}}@{\hspace{0.5em}}c@{\hspace{0.5em}}@{\hspace{0.5em}}c@{\hspace{0.5em}}@{\hspace{0.5em}}c@{\hspace{0.5em}}@{\hspace{0.5em}}c@{\hspace{0.5em}}@{\hspace{0.5em}}c@{\hspace{0.5em}}@{\hspace{0.5em}}c@{\hspace{0.5em}}@{\hspace{0.5em}}c@{\hspace{0.5em}}}
    \toprule
    & \multicolumn{4}{c}{German}&& \multicolumn{4}{c}{French} && \multicolumn{4}{c}{Japanese} \\
    \cmidrule{2-5}\cmidrule{7-10}\cmidrule{12-15}
    Domain & books & dvd & music & \emph{avg} && books & dvd & music & \emph{avg} && books & dvd & music & \emph{avg} \\
    \midrule
    \multicolumn{15}{l}{\emph{Methods with general-purpose cross-lingual resources}}\\
    MT-BOW\tnote{1}  & 79.68 & 77.92 & 77.22 & 78.27 &  & 80.76 & 78.83 & 75.78 & 78.46 &  & 70.22 & 71.30 & 72.02 & 71.18  \\
    CL-SCL\tnote{1}  & 79.50 & 76.92 & 77.79 & 78.07 &  & 78.49 & 78.80 & 77.92 & 78.40 &  & \underline{73.09} & 71.07 & 75.11 & 73.09  \\
    CR-RL\tnote{2}  & 79.89 & 77.14 & 77.27 & 78.10 &  & 78.25 & 74.83 & 78.71 & 77.26 &  & 71.11 & 73.12 & 74.38 & 72.87  \\
    Bi-PV\tnote{3}  & 79.51 & 78.60 & \bf \underline{82.45} & 80.19 &  & \bf \underline{84.25} & 79.60 & \underline{80.09} & \underline{81.31} &  & 71.75 & \underline{75.40} & \underline{75.45} & \underline{74.20}  \\
    UMM\tnote{4}  & \underline{81.65} & \underline{81.27} & 81.32 & \underline{81.41} &  & 80.27 & \underline{80.27} & 79.41 & 79.98 &  & 71.23 & 72.55 & 75.38 & 73.05 \\
    \midrule
    \multicolumn{15}{l}{\emph{Methods with task-specific cross-lingual resources}}\\
    CLDFA\tnote{5}  & \bf \underline{83.95} & \bf \underline{83.14} & \underline{79.02} & \bf \underline{82.04} &  & \underline{83.37} & \underline{82.56} & \bf \underline{83.31} & \bf \underline{83.08} &  & \bf \underline{77.36} & \bf \underline{80.52} & \bf \underline{76.46} & \bf \underline{78.11}  \\
    \midrule
    \multicolumn{15}{l}{\emph{Methods without cross-lingual resources}}\\
    BWE (1-to-1)  & 76.00 & 76.30 & 73.50 & 75.27 &  & 77.80 & 78.60 & 78.10 & 78.17 &  & 55.93 & 57.55 & 54.35 & 55.94 \\
    BWE (3-to-1)  & 78.35 & 77.45 & 76.70 & 77.50 &  & 77.95 & 79.25 & 79.95 & 79.05 &  & 54.78 & 54.20 & 51.30 & 53.43 \\
    \manmoe{}  & \underline{82.40} & \underline{78.80} & \underline{77.15} & \underline{79.45} &  & \underline{81.10} & \bf \underline{84.25} & \underline{80.90} & \underline{82.08} &  & \underline{62.78} & \underline{69.10} & \underline{72.60} & \underline{68.16} \\
    \bottomrule
\end{tabular}
\begin{tablenotes}
\noindent\item[1] \citet{P10-1114}
\noindent\item[2] \citet{D13-1153}
\noindent\item[3] \citet{pham-luong-manning:2015:VSM-NLP}

\noindent\item[4] \citet{xu-wan:2017:EMNLP2017}
\noindent\item[5] \citet{P17-1130} 
\end{tablenotes}
\end{threeparttable}
\caption{Results for the Multilingual Amazon Reviews dataset. Numbers indicate binary classification accuracy.
VecMap embeddings~\citep{P17-1042} are used for this experiment as MUSE training fails on Japanese (\secref{sec:model_arch}).}
\label{tab:amazon_results}
\vspace{-3mm}
\end{table*}

In a contemporaneous work by~\citep{D18-1034}, they propose a cross-lingual NER model using Bi-LSTM-CRF that achieves similar performance compared to \manmoe{}+CharCNN.
However, our architecture is not specialized to the NER task, and we did not add task-specific modules such as a CRF decoding layer, etc.

Last but not least, we replace the MUSE embeddings with the recently proposed unsupervised multilingual word embeddings~\citep{chen2018umwe}, which further boosts the performance, achieving a new state-of-the-art performance as shown in Table~\ref{tab:ner_results} (last row).

\subsection{Cross-Lingual Text Classification on Amazon Reviews}\label{sec:exp:amazon}

\begin{figure}
    \centering
    \includegraphics[width=\linewidth]{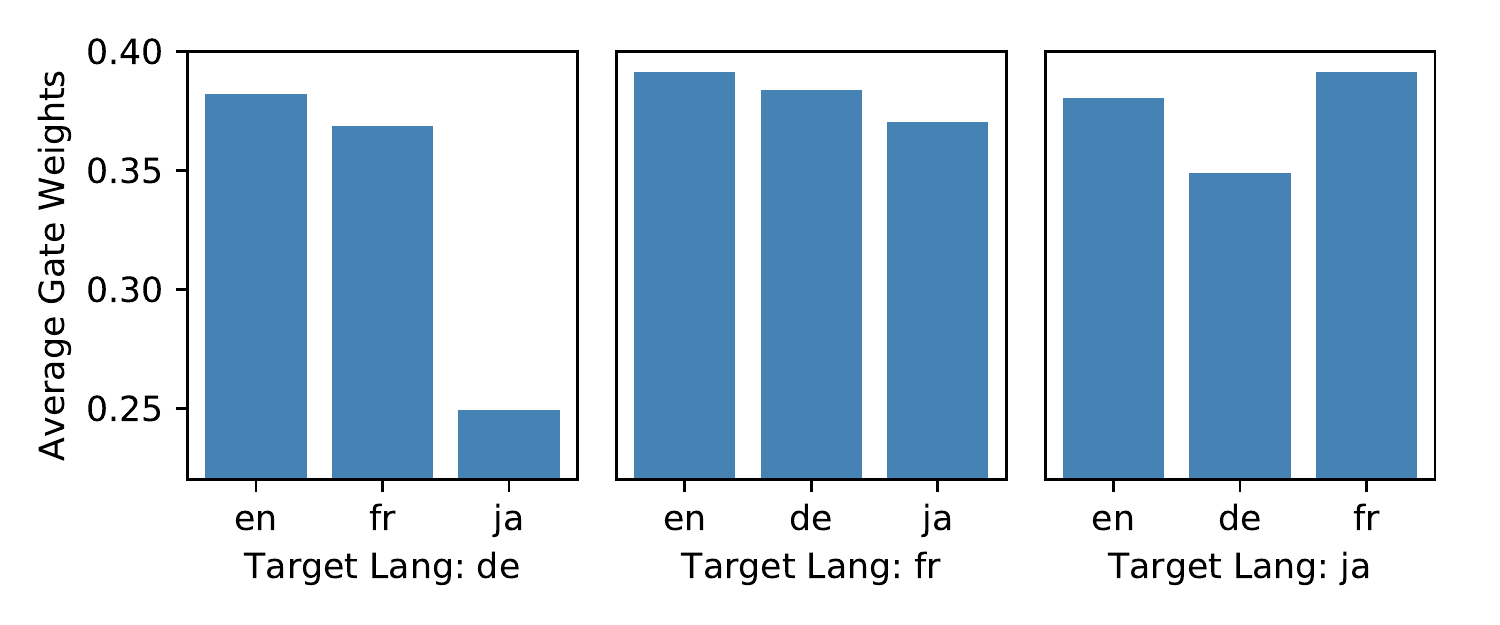}
    \caption{Average expert gate weights aggregated on a language level for the \emph{Amazon Reviews} dataset.}
    \vspace{-3mm}
    \label{fig:avg_gate_weights}
\end{figure}

Finally, we report results on a multilingual text classification dataset~\cite{P10-1114}. 
The dataset is a binary classification dataset where each review is classified into positive or negative sentiment.
It has four languages: English, German, French and Japanese.

As shown in Table~\ref{tab:amazon_results}, MT-BOW uses machine translation to translate the bag of words of a target sentence into the source language, while CL-SCL learns a cross-lingual feature space via structural correspondence learning~\cite{P10-1114}.
CR-RL~\citep{D13-1153} learns bilingual word representations where part of the word vector is shared among languages.
Bi-PV~\citep{pham-luong-manning:2015:VSM-NLP} extracts bilingual paragraph vector by sharing the representation between parallel documents.
UMM~\citep{xu-wan:2017:EMNLP2017} is a multilingual framework that could utilize parallel corpora between multiple language pairs, and pivot as needed when direct bitexts are not available for a specific source-target pair.
Finally CLDFA~\citep{P17-1130} proposes cross-lingual distillation on parallel corpora for CLTL.
Unlike other works listed, however, they adopt a task-specific parallel corpus (translated Amazon reviews) that are difficult to obtain in practice, making the numbers not directly comparable to others.

Among these methods, UMM is the only one that does not require direct parallel corpus between all source-target pairs.
It can instead utilize pivot languages (e.g.~English) to connect multiple languages.
\manmoe{}, however, takes another giant leap forward to completely remove the necessity of parallel corpora while achieving similar results on German and French compared to UMM.
On Japanese, the performance of \manmoe{} is again limited by the quality of BWEs. (BWE baselines are merely better than randomness.)
Nevertheless, \manmoe{} remains highly effective and the performance is only a few points below most SoTA methods with cross-lingual supervision.

For a better understanding of the model behavior, Figure~\ref{fig:avg_gate_weights} visualizes the expert weights when transferring to different languages, which corroborates our model hypothesis and the findings in \secref{sec:exp:ablation} (see Appendix~\ref{sec:visualization} for more details).

\section{Conclusion}\label{sec:conclusion}

In this paper, we propose \manmoe{}, a multilingual model transfer approach that exploits both language-invariant (shared) features and language-specific (private) features, which departs from most previous models that can only make use of shared features.
Following earlier work, the shared features are learned via language-adversarial training~\citep{chen2016adan}.
On the other hand, the private features are extracted by a mixture-of-experts (\moe{}) module, which is able to dynamically capture the relation between the target language and each source language on a token level.
This is extremely helpful when the target language is similar to a subset of source languages, in which case traditional models that solely rely on shared features would perform poorly.
Furthermore, \manmoe{} is a purely model-based transfer method, which does not require parallel data for training, enabling fully zero-resource MLTL when combined with unsupervised cross-lingual word embeddings.
This makes \manmoe{} more widely applicable to lower-resourced languages.

Our claim is supported by a wide range of experiments over multiple text classification and sequence tagging tasks, including a large-scale industry dataset.
\manmoe{} significantly outperforms all cross-lingually unsupervised baselines regardless of task or language.
Furthermore, even considering methods with strong cross-lingual supervision, \manmoe{} is able to match or outperform these models on closer language pairs.
When transferring to distant languages such as Chinese or Japanese (from European languages), where the quality of cross-lingual word embeddings are unsatisfactory, \manmoe{} remains highly effective and substantially mitigates the performance gap introduced by cross-lingual supervision.

For future work, we plan to apply \manmoe{} to more challenging languages for tasks such as syntactic parsing, where multilingual data exists~\cite{11234/1-2184}.
Furthermore, we would like to experiment with multilingual contextualized embeddings such as the Multilingual BERT~\cite{devlin2018bert}.

\bibliography{manmoe}
\bibliographystyle{acl_natbib}

\clearpage

\appendices
\begin{figure*}
    \centering
    \includegraphics[width=0.9\linewidth]{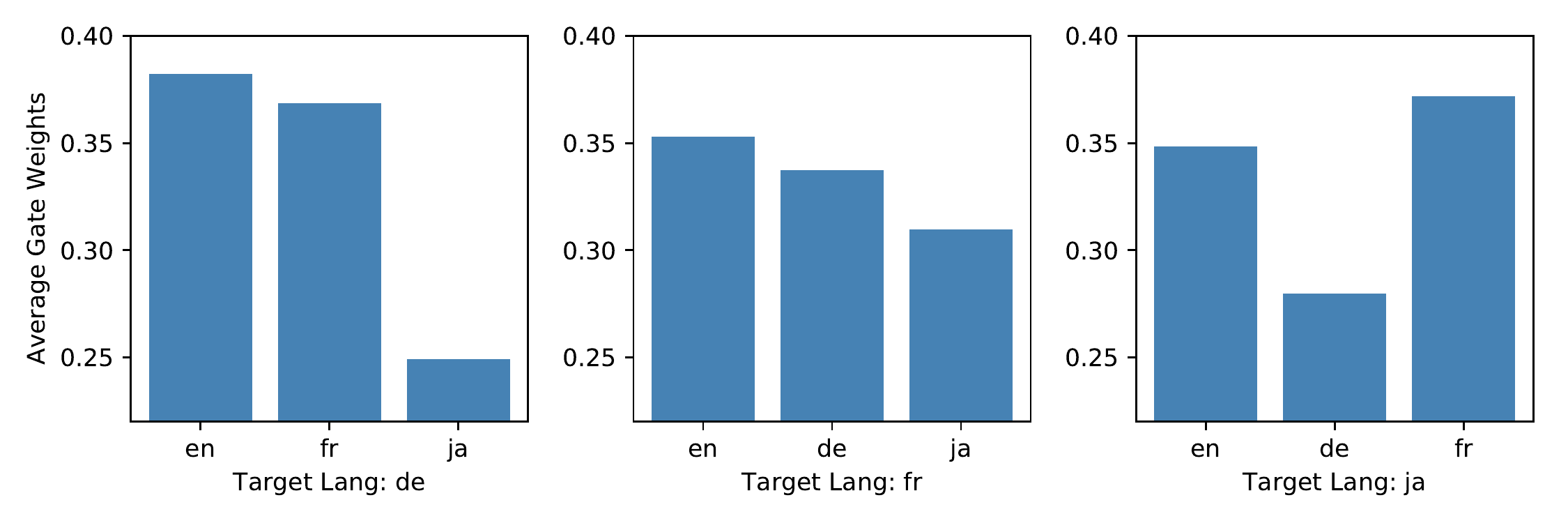}
    \caption{Average expert gate weights aggregated on a language level for the Amazon dataset.}
    \label{fig:avg_gate_weights_supl}
\end{figure*}

\begin{figure*}
    \centering
    \includegraphics[width=0.9\linewidth]{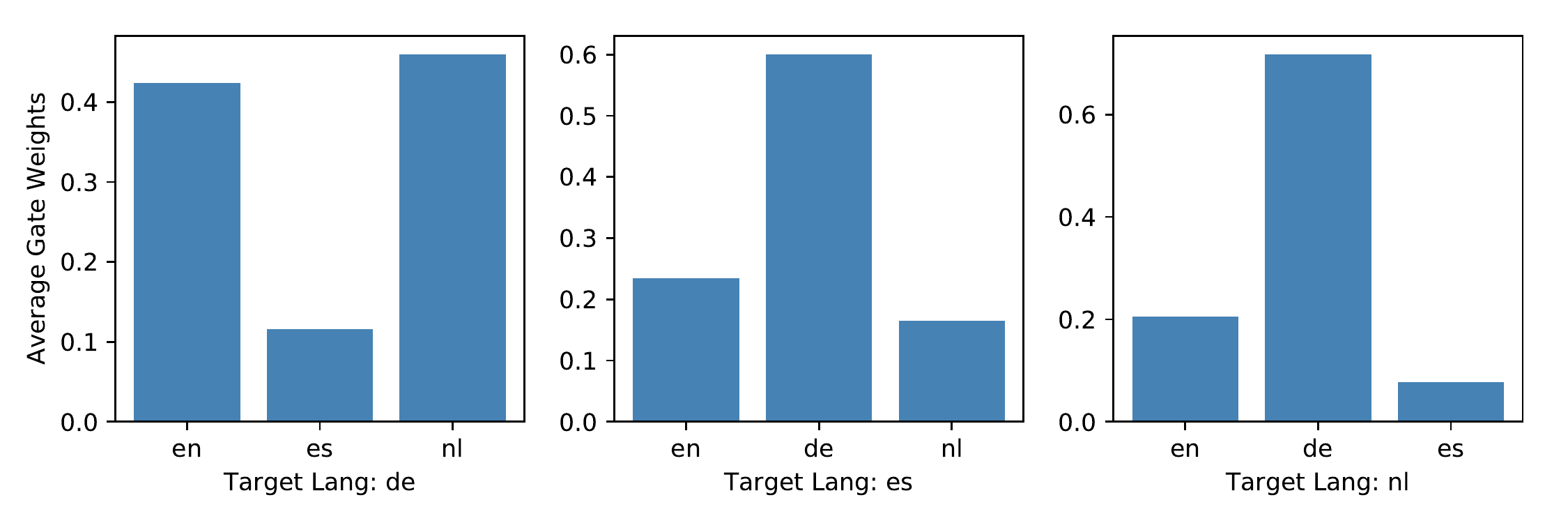}
    \caption{Average expert gate weights aggregated on a language level for the CoNLL dataset.}
    \label{fig:avg_gate_weights_supl_ner}
\end{figure*}

\section{Visualization of Expert Gate Weights}\label{sec:visualization}

In Figure~\ref{fig:avg_gate_weights_supl} and~\ref{fig:avg_gate_weights_supl_ner}, we visualize the average expert gate weights for each of the three target languages in the Amazon and CoNLL datasets, respectively.
For each sample, we first compute a sentence-level aggregation by averaging over the expert gate weights of all its tokens.
These sentence-level expert gate weights are then further averaged across all samples in the validation set, which forms a final language-level average expert gate weight for each target language.
For the Amazon dataset, we take the combination of all three domains (books, dvd, music).

The visualization further collaborates with our hypothesis that our model makes informed decisions when selecting what features to share to the target language.
On the Amazon dataset, it can be seen that when transferring to German or French (from the remaining three), the Japanese expert is less utilized compared to the European languages.
On the other hand, it is interesting that when transferring to Japanese, the French and English experts are used more than the German one, and the exact reason remains to be investigated.
However, this phenomenon might be of less significance since the private features may not play a very important role when transferring to Japanese as the model is probably focusing more on the shared features, according to the ablation study in Section~\ref{sec:exp:ablation}.

In addition, on the CoNLL dataset, we observe that when transferring to German, the experts from the two more similar lanaguages, English and Dutch, are favored over the Spanish one.
Similarly, when transferring to Dutch, the highly relevant German expert is heavily used, and the Spanish expert is barely used at all.
Interestingly, when transferring to Spanish, the model also shows a skewed pattern in terms of expert usage, and prefers the German expert over the other two.
\section{Implementation Details}\label{sec:implementation}

\begin{figure}
    \centering
    \includegraphics[width=0.8\linewidth]{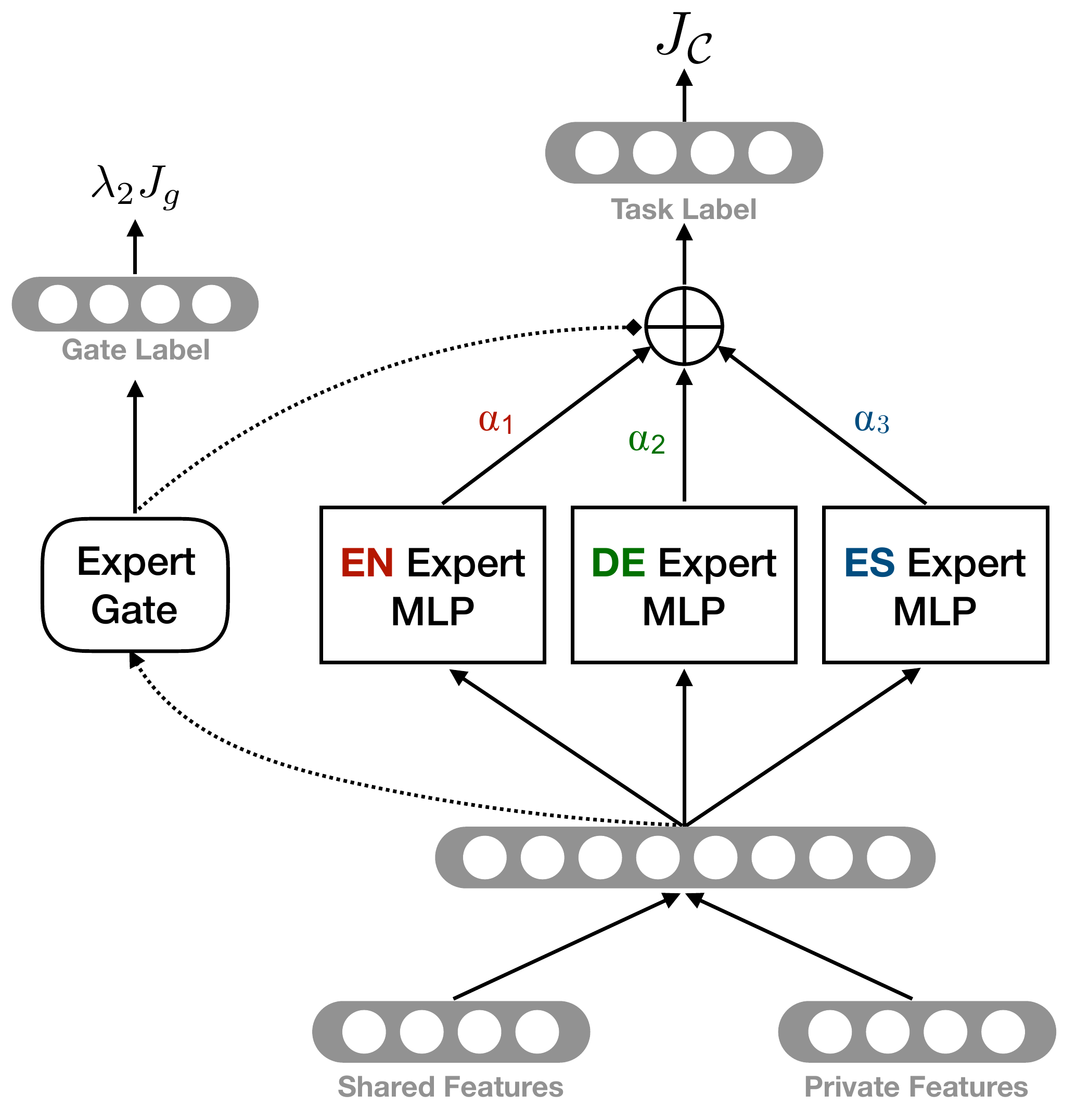}
    \caption{The \moe{} Predictor $\gC$ for Sequence Tagging.}
    \label{fig:c}
\end{figure}

In all experiments, Adam~\cite{kingma2014adam} is used for both optimizers (main optimizer and $\gD$ optimizer), with learning rate $0.001$ and weight decay $10^{-8}$.
Batch size is 64 for the slot filling experiment and 16 for the NER and Amazon Reviews experiments, which is selected mainly due to memory concerns.
CharCNN increases the GPU memory usage and NER hence could only use a batch size of 16 to fit in 12GB of GPU memory.
The Amazon experiment does not employ character embeddings but the documents are much longer, and thus also using a smaller batch size.
All embeddings are fixed during training.
Dropout~\cite{JMLR:v15:srivastava14a} with $p=0.5$ is applied in all components.
Unless otherwise mentioned, ReLU is used as non-linear activation.

Bidirectional LSTM is used in the feature extractors for all experiments.
In particular, $\gF_s$ is a two-layer BiLSTM of hidden size 128 (64 for each direction), and $\gF_p$ is a two-layer BiLSTM of hidden size 128 stacked with a \moe{} module (see Figure~\ref{fig:fp}).
Each expert network in the \moe{} module of $\gF_p$ is a two-layer MLP again of hidden size of 128.
The final layer in the MLP has a $tanh$ activation instead of ReLU to match the LSTM-extracted shared features (with $tanh$ activations).
The expert gate is a linear transformation (matrix) of size $128\times N$, where $N$ is the number of source languages.

On the other hand, the architecture of the task specific predictor $\gC$ depends on the task.
For sequence tagging experiments, the structure of $\gC$ is shown in Figure~\ref{fig:c}, where each expert in the \moe{} module is a token-level two-layer MLP with a softmax layer on top for making token label predictions.
For text classification tasks, a dot-product attention mechanism~\cite{D15-1166} is added after the shared and private features are concatenated.
It has a length 256 weight vector that attends to the feature vectors of each token and computes a softmax mixture that pools the token-level feature vectors into a single sentence-level feature vector.
The rest of $\gC$ remains the same for text classification.

\begin{table}
    \centering
    \begin{tabular}{l c c c}
        \toprule
        & $\lambda_1$ & $\lambda_2$ & $k$ \\
        \midrule
        Slot Filling    & $0.01$ & $1$ & $5$ \\
        CoNLL NER & $0.0001$ & $0.01$ & $1$ \\
        Amazon & $0.002$ & $0.1$ & $1$ \\
        \bottomrule
    \end{tabular}
    \caption{The hyperparameter choices for different experiments.}
    \label{tab:hyperparam}
\end{table}

For the language discriminator $\gD$, a CNN text classifier~\cite{D14-1181} is adopted in all experiments.
It takes as input the shared feature vectors of each token, and employs a CNN with max-pooling to pool them into a single fixed-length feature vector, which is then fed into a MLP for classifying the language of the input sequence.
The number of kernels is 200 in the CNN, while the kernel sizes are 3, 4, and 5.
The MLP has one hidden layer of size 128.

The MUSE, VecMap, and UMWE embeddings are trained with the monolingual $300d$ fastText Wikipedia embeddings~\citep{bojanowski2016enriching}.
When character-level word embeddings are used, a CharCNN is added that takes randomly initialized character embeddings of each character in a word, and passes them through a CNN with kernel number 200 and kernel sizes 3, 4, and 5.
Finally, the character embeddings are max-pooled and fed into a single fully-connected layer to form a 128 dimensional character-level word embedding, which is concatenated with the pre-trained cross-lingual word embedding to form the final word representation of that word.

The remaining hyperparameters such as $\lambda_1$, $\lambda_2$ and $k$ (see Algorithm~\ref{alg:training}) are tuned for each individual experiment, as shown in Table~\ref{tab:hyperparam}.

\end{document}